# Applying Transfer Learning for Improving Domain-Specific Search Experience Using Query to Question Similarity

Ankush Chopra
Artificial Intelligence, CoE
Fidelity Investments
Bengaluru, Karnataka, India
ankush01729@gmail.com

Shruti Agrawal
Artificial Intelligence, CoE
Fidelity Investments
Bengaluru, Karnataka, India
shruti.agrawal271@gmail.com

Sohom Ghosh
Artificial Intelligence, CoE
Fidelity Investments
Bengaluru, Karnataka, India
sohom1ghosh@gmail.com

## ABSTRACT

Search is one of the most common platforms used to seek information. However, users mostly get overloaded with results whenever they use such a platform to resolve their queries. Nowadays, direct answers to queries are being provided as a part of the search experience. The question-answer (QA) retrieval process plays a significant role in enriching the search experience. Most off-the-shelf Semantic Textual Similarity models work fine for well-formed search queries, but their performances degrade when applied to a domain-specific setting having incomplete or grammatically ill-formed search queries in prevalence. In this paper, we discuss a framework for calculating similarities between a given input query and a set of predefined questions to retrieve the question which matches to it the most. We have used it for the financial domain, but the framework is generalized for any domain-specific search engine and can be used in other domains as well. We use Siamese network [6] over Long Short-Term Memory (LSTM) [3] models to train a classifier which generates un-normalized and normalized similarity scores for a given pair of questions. Moreover, for each of these question pairs, we calculate three other similarity scores: cosine similarity between their average word2vec embeddings [15], cosine similarity between their sentence embeddings [7] generated using RoBERTa [17] and their customized fuzzy-match score. Finally, we develop a meta-classifier using Support Vector Machines [19] for combining these five scores to detect if a given pair of questions is similar. We benchmark our model's performance against existing State Of The Art (SOTA) models on Quora Question Pairs (QQP) dataset[1] as well as a dataset specific to the financial domain. After evaluating its performance on the financial domain specific data, we conclude that it not only outperforms several existing SOTA models on F1 score but also has decent accuracy.

## 1 INTRODUCTION

Enterprise search platforms usually cater to the needs of users who look for information relating to a certain domain. A lot of work relating to text-similarity has been done in recent times. However, most of these works are generic. Let's think of an instance where a user searches for "poa". What he means, in the financial domain is "What is power of attorney?". The generic models do not perform well in these kinds of scenarios. They are not able to effectively measure the similarity between search queries and underlying content. For a given domain, the search space is limited. In general, novel search queries are not seen very often. However, there is a lot of variability in how the same information need is expressed in the form of actual search queries. For instance, "how to open an Individual Retirement Acc.?" and "Tell me the steps that I need to follow for opening an IRA account" carry the same sense but are presented differently. Languages are ever evolving. They are the means to express one's thoughts and emotions. Thus, they are used very creatively. Users looking for answers to their queries use variations of sentences which are semantically similar, but they may be syntactically different (such as short and misspelt queries). We already have answers for the queries which are frequently searched. A model capable of mapping a new variation of a question to those in the pre-existing question corpus enables us in providing direct answers to users' queries. For developing such a model, a lot of domain-specific data is needed. Due to the unavailability of such a huge corpus, we use transfer learning.

Firstly, we use the publicly available QQP dataset[1] to build a baseline model and then fine-tune it further using the limited domain-specific data that we have. The data mentioned latter consists of the logs of searches and chats which were done on Fidelity.com. This data relates to the financial domain.

In this model, we use Siamese [6] LSTM [3] architecture which takes two sentences as input and detects if they are similar. It uses readily available 300-dimensional word2vec embedding [15] trained on Google News corpus to represent the inputs numerically. Its final layer learns cosine similarities between vector representations of the sentences. While implementing this, we realize that users do not always write search queries in the form of questions. Moreover, most of them are not grammatically correct. Analyzing the instances which were misclassified by this model, we discover that short, mistyped and grammatically incorrect queries degrade its performance. Thus, it is essential to come up with a framework which is capable of dealing with these shortcomings present in the search queries of the domain-specific data. To overcome these shortcomings, we propose a model which ensembles normalized and un-normalized scores generated by the baseline model along with a novel fuzzy-match score and a cosine similarity score between their average word2vec embedding [15].

Recent advancements in NLP using deep learning have resulted in several pre-trained transformer-based embedding models like BERT [11]. These models outperform existing embedding-based models like word2vec [15] and GloVe [20] in most of the NLP tasks. Thus, as an enhancement, we perform several experiments using cosine similarities between sentence embedding [7] of input queries generated using XLNet [1], BERT [11], DistillBERT [16] and RoBERTa [17]. We observe that by including RoBERTa, the precision of the final classification model is the best for our domain-specific dataset.

Since this model uses five similarity scores, it would take a lot of time to scan through the entire set of curated questions for finding out that question which matches the most to the input query. Thus, we have also come up with an efficient retrieval system that can return the best-matched question for any incoming customer query in real-time.

Our main contributions are as follows:
- Proving that off-the-shelf SOTA models do not perform well on domain specific data
- Presenting a novel architecture for measuring similarity between search query and questions
- Using transfer learning to enhance its performance on data specific to the financial domain
- Successfully deploying it in production for providing direct answers to searches done by users in real-time

This paper is structured into seven major sections. This section introduced readers to the journey which led us to come up with this novel framework. This framework improves users' search experience using semantic and syntactic textual similarity. In the next section, we extensively explore the related frameworks which are already existing. The third section states our problem statement formally. It also familiarizes everyone with the datasets being used. We narrate the methodologies used to develop the new framework/model in the fourth section. In the fifth section, we elucidate several experiments which we have conducted and their results. The section after that states how we have effectively put this into production. We conclude and reveal our plans to improve this model in the last section.

## 2 RELATED WORKS

This section discusses the works which are relevant for us and have been done previously along with their performances on standard semantic textual similarity tasks. Siamese Recurrent Neural architectures have been used by Mueller et. al in their paper [6]. They propose Manhattan LSTM Model which gives Pearson correlation of 0.8822 and Spearman's correlation of 0.8345 on SICK semantic textual similarity task [9]. In [2], Chali et al. describe how they re-define existing SOTA by using deep learning-based architectures like LSTM [3], LSTM with attention [10], Bi-directional LSTM stacking with attention and stacked Bi-directional LSTM with attention which includes a Convoluted Neural Network (CNN) with max-pooling over it. The latter one provides the best accuracy i.e. 88.8% while detecting duplicate questions in the QQP dataset[1]. Their work specifically focuses on online forums. Tan et al. in their paper [5] illustrate how they use Multi-way Attention Networks for developing a classifier capable of detecting duplicate questions, identifying paraphrases, inferring natural language and selecting an answer from a given text corpus. Yang et al. [14] propose a hierarchical attention network for representing text documents. It consists of two parts: word encoder followed by word attention. The second part consists of sentence encoder followed by sentence attention. Although they focus mainly on classification, their method of combining word vectors is quite innovative. Wang et al. in their paper [13] have shown how similarity between texts can be calculated using lexical decomposition and composition. Their model consists of 5 parts: word representation, semantic matching, decomposition, composition and similarity assessing. Its performance is better for the task of question answering than for identification of paraphrases. Wang et al. in [12] deal with the Natural Language Sentence Matching (NLSM) task in two phases. Firstly, they develop a model consisting of 5 layers: word representation layer, context representation layer, matching layer, aggregation layer and prediction layer. Their main novelty lies in the second phase where they propose four different matching perspectives: full matching, max-pooling matching, attentive matching and max-attentive matching. They evaluate this bilateral multi-perspective matching model (BiMPM) on three different NLSM tasks namely answer selection for questions, natural language inference and paraphrase identification. BiMPM's performance redefines the state-of-the-art benchmark in detecting paraphrase on the QQP dataset[1] and inferring natural language on the SNLI dataset[2]. Devlin et. al in their paper Bidirectional Encoder Representations from Transformers (BERT) [11] introduce a novel concept of masking to create language models and elucidate how they do the bidirectional training using Transformers [10]. This model for detecting duplicate questions, achieves an accuracy of 89.30% and F1 score (un-weighted, macro average) of 0.721 on QQP dataset[1] as per the GLUE[3] [21] leaderboard. Two variants of BERT [11] namely RoBERTa [17] and DistilBERT [16] were proposed by Liu et al. and Sanh et al. respectively. In the paper RoBERTa [17], Liu et al. narrate how they change the masking pattern of BERT [11] dynamically, remove the objective of the next sentence prediction and train it for more time on longer text sequences. It gives an accuracy of 90.20% and F1 score of 0.743 on QQP[1] dataset as mentioned in the GLUE[3] [21] leaderboard. The DistilBERT [16] model uses knowledge distillation process while pre-training. They use three kinds of loss functions: cosine-distance loss, the triple loss (which occurs when language models are combined) and distillation loss. On QQP dataset[1], they achieve an accuracy of 88.5%. Yang et al. in their paper [1], propose XLNet which is based on [8] Transformer-XL. It uses auto-regressive methods and auto-encoding methods with the objective of permutation language modelling. It overcomes the shortcomings of BERT [11] model like neglecting dependency between masked positions. As per the GLUE[3] [21] leaderboard, the ensemble of XLNet models for detecting duplicate questions achieves an accuracy of 90.4% and F1 score of 0.747 on QQP[1] dataset. In [4] Sakata et al. discuss how they use question-question similarity (by TSUBAKI) and question-answer relevance using BERT [11] to retrieve Frequently Asked Questions effectively. Reimers et al. in [7] discuss Sentence-BERT. They create it by modifying siamese architecture and adding a triplet network with pre-trained BERT. They experiment with three kinds of objective functions: classification, regression and triplet. They perform the evaluation on Semantic Text Similarity tasks and SentEval transfer tasks. This Sentence-BERT outperforms existing ways of creating sentence embedding. However, none of the models described here is fine-tuned for search queries specific to a given domain like finance.

---

[1] https://data.quora.com/First-Quora-Dataset-Release-Question-Pairs

[2] https://nlp.stanford.edu/projects/snli/

## 3 PROBLEM STATEMENT

### 3.1 Task Definition

For a given input query r and a pre-existing curated set of questions $\{q_1, q_2..., q_n, q_{n+1}\}$, we need to develop an algorithm which selects one question $q \in \{q_1, q_2...,q_n, q_{n+1}\}$ such that the similarity between r and q is the maximum. The supervised learning-based re-formulation of the problem is as follows:
Given a set of question-question pairs and their corresponding labels $\{(q_1, q_2, l_{(1,2)}), (q_3, q_4, l_{(3,4)}) ... (q_n, q_{n+1}, l_{(n,n+1)})\}$ where each label $l_{(i,j)} \in \{0,1\}$ denotes whether $q_i$ and $q_j$ is identical (when equal to 1) or not (when equal to 0). We develop a model which is capable of detecting if a given question pair $(q_x, q_y)$ means the same. It also returns a score which states the similarity between $q_x$ and $q_y$.

### 3.2 Understanding the data set

We have primarily used two datasets. The first one consists of the QQP dataset[1] which is available publicly. It has 404,290 question-question pairs out of which 149,263 instances are duplicate. It has total 537,361 distinct questions. Out of the entire dataset, 323,479 question pairs have been used for training the initial Siamese [6] LSTM [3] model. Its event rate is 37%.

The other dataset is internal to our organization, Fidelity Investments. It is from the financial domain. It comprises of queries frequently asked or searched by users. Furthermore, questions raised by them while chatting with customer representatives are also used. 6,391 question pairs are randomly picked. Each of them is assigned a label manually. It has a total of 6,419 unique questions. It is used for initial training of this Siamese [6] LSTM [3] model as well as for the training of Support Vector Machine based meta-classifier [19]. Out of these 6,391 instances, 5,196 instances are used for training and 1,195 instances are used for validation. The event rate in each case is 67%. There is no instance which is common to both of these datasets.

## 4 METHODOLOGIES

We initiate by describing an important component of the proposed ensemble model, the Siamese [6] LSTM [3] network. LSTM [3] models are a special case of Recurrent Neural Networks (RNNs). They can remember the things which are essential and forget the rest. This makes them more useful and provides a longer memory. There are two LSTM [3] networks to process one question each. Siamese [6] network is a type of neural network where weights of the network are shared between different intermediate layers. This can be understood as more than one sub-network which have the same configurations and weights of parameters. As the networks' weights for both the inputs are same, this network performs well on the tasks like the comparison of texts or finding similarity between input vectors. These input vectors are created using word2vec [15]. It takes two questions as input and returns their similarity score. It is tuned to produce better precision of the positive class. Its hyperparameters are mentioned in Figure 1. We use this component to generate the un-normalized and normalized scores. We also calculate cosine similarity scores using the average embedding of input texts. Furthermore, we use customized fuzzy-match score and cosine similarity based on sentence embedding [7] generated using RoBERTa [17]. We describe each of these five scores in detail in the subsequent parts of this section. For combining these scores, we use a support vector machine [19] based meta-classifier. It is elaborated at the end of this section. The proposed architecture consists of the following 5 features (similarity scores):

### 4.1 Un-normalized score

This similarity score is computed using the Siamese model trained on un-normalized question pairs. Some pre-processing steps are involved before sending the input to the Siamese [6] LSTM [3] architecture to compute the un-normalized score. These pre-processing steps are as follows:
**Basic cleaning:** This includes common pre-processing steps like converting to lower case, removal of extra spaces, removal of punctuation marks and other regex-based replacements.
**Contractions replacements:** This replaces the contractions with their expanded form. E.g. "I've" gets changed to "I have" and "hadn't've" gets replaced by "had not have".

Apart from the above-mentioned steps, we also did the following pre-processing which is specific to our internal dataset from the financial domain:

---

[3] https://gluebenchmark.com/leaderboard (as on 28th September 2020)

**Product variation normalization:** Variations of common product entities get normalized into one token using this module. This normalization happens using a predefined mapping of the variations of different product and service names. E.g. 401(K) gets mapped to 401k.

**Acronym identification and replacement:** It helps in identifying the presence of an acronym and replaces it with its appropriate expansion. It also uses a predefined mapping of acronyms and their definitions. E.g. "What is an IRA?", "what is an Individual retirement account (IRA)?" and "What are IRAs?" are treated the same and they become "What is/are individual retirement account?".

```
Layer (type)                 Output Shape        Param #     Connected to
==================================================================================
input_1 (InputLayer)         (None, 144)         0
input_2 (InputLayer)         (None, 144)         0
embedding_1 (Embedding)      (None, 144, 300)    26808300    input_1[0][0]
                                                             input_2[0][0]
lstm_1 (LSTM)                (None, 75)          112800      embedding_1[0][0]
                                                             embedding_1[1][0]
dropout_1 (Dropout)          (None, 75)          0           lstm_1[0][0]
dropout_2 (Dropout)          (None, 75)          0           lstm_1[1][0]
dot_1 (Dot)                  (None, 1)           0           dropout_1[0][0]
                                                             dropout_2[0][0]
==================================================================================
```

Figure 1: Siamese LSTM Model Hyper-parameters

### 4.2 Normalized score

This similarity score is computed using the Siamese model on normalized question pairs. The normalized score generation process includes the following pre-processing steps in addition to the steps mentioned above:

**Lemmatization:** It uses Spacy's[4] Part of Speech (PoS) tagger and converts all the words which are identified as verbs to their base forms.

**Singularization:** It converts plural common nouns to their singular form. Spacy[4] PoS tagger is used to identify the candidate words to be passed through this submodule.

The above normalization of the query helps in sending the de-noised and base forms of possible variations of the inputs to the model. We keep the un-normalized score along with this to make sure that we don't lose any useful information and let the ensemble model assign a weight to each of these.

### 4.3 Average embedding similarity score

This score is useful in the cases where the input query is grammatically incorrect. Users often tend to write incomplete search queries making it difficult for models to identify meaningful features related to sentence structure. For example: "eft routing number" and "what is the routing number for electronic fund transfer" don't seem similar to the model. We get the embedding for each word from the embedding matrix of the Siamese [6] LSTM [3] model. The 300-dimensional embedding vector for each word of the sentence is added and divided by the total number of words to get the resultant average embedding vector representation for the sentence. The average embedding based similarity score is then calculated using the dot product of the input query vector with that of each question in the QA corpus.

### 4.4 Customized fuzzy-match score

Often users performing search inadvertently misspell words. The deep learning model we trained takes care of semantic matching between the sentences. However, it fails when misspelt words occur because it treats such words as words which are not present in its vocabulary. This brings in the need for a fuzzy-match score. This score helps in taking care of misspelt query inputs and returns the similarity score that represents the extent to which a query matches a question syntactically. Details of this novel method are mentioned in Algorithm 1. This algorithm demonstrates the computation of the customized fuzzy-match score for 2 sentences. We tokenize the sentences after basic cleaning and replacements and check for overlap/intersection of tokens between them. If there is no overlap, the custom fuzzy-match score is 0. In cases with overlap, for each word exclusive to sentence 1, we check how similar is it with words exclusive to sentence 2. To calculate this similarity, we follow 2 ways:

**Syntactic match:** This uses the normalized Levenshtein distance which ranges from 0 to 1 to compute the syntactic difference between a given pair of words. The syntactic match score (1 − normalized Levenshtein distance) is compared to the pre-defined threshold (0.6). If the score fails to cross the threshold, implying that the two words are not syntactically similar, then we move on to the next step to check semantic similarity.

**Semantic match:** We compute the cosine similarity between the two word vectors and scale it to get the semantic score between 0 to 1. This is compared with a threshold (0.55) which is estimated empirically.

As both the scores denote the similarity scaled between 0 to 1, these are added and used further in computation of the final custom score. The final score is the ratio between the total overlap/similarity between the tokens and the total number of unique meaning tokens considering both the sentences. The thresholds for the syntactic and semantic score were defined after experimentation with a grid of values on a small subset of training data.

For example, a customer types "What is cost for factonal trading?" and we have a question in QA bank "What are fees or charges for fractional trading?". After basic pre-processing, words exclusive to sentence1 are ["cost", "factnol"] and sentence2 are ["fees", "charges", "fractional"]. The algorithm would first check for the syntactic score of ("cost", "fees"). As this would fail to cross the pre-defined threshold, it would then compute the semantic score of the pair. Since the semantic score would cross the threshold to imply that the words have a similar meaning, we would then move on to the next word pair and so on. The customized fuzzy-match score will thus take care of pairs ("cost","charges") and ("cost","fees") using its semantic score and it would take care of ("factnol","fractional") using the syntactic score.

Any off-the-shelf syntactic match model would return a lower score for the above two sentences considering the syntactic difference between "fees"/"charges" and "cost". But, this custom fuzzy-match score takes care of any such instances by bringing in the semantic similarity feature as well. The novelty of the algorithm lies in the fact that it uses word2vec [15] similarity to calculate custom fuzzy scores and thus, performs better than off-the-shelf syntactic similarity models.

---

[4] https://spacy.io/

## 4.5 RoBERTa based score

Recently, there have been major advancements in the field of NLP. Using pre-trained embeddings is the new normal. We've many BERT [11] like Transformer [10] based models that outperform the existing embeddings like word2vec [15] and GloVe [20] for many NLP tasks like question answering, text summarization and so on. So, as an enhancement, we include cosine similarity between sentence embedding [7] of questions generated using RoBERTa [17] large model as one more feature to the meta-classifier. This universal sentence embedding is created by using mean token pooling. It has been trained on SNLI[2] and MultiNLI[5] dataset. By including this score, the accuracy improves on the internal test set.

To combine these scores, we trained a support vector machine [19] based meta-classifier model. This is outlined in Figure 2. The hyper-parameters of this model, producing the best output are: degree = 2, kernel type = polynomial and regularization (C) = 0.2.

To summarize, the proposed architecture computes the similarity between a search query and the questions present in a QA corpus by providing these five scores as inputs to the meta-classifier (SVC).

## 5 EXPERIMENTATION AND RESULTS

In this section, we describe various experiments which we have done and their results. For representing input texts numerically, we use word2vec [15] embedding. We train a Siamese [6] LSTM [3] network to compute the similarity score between questions using the QQP dataset[1]. We tune it to produce the best precision for the positive class. While evaluating this Siamese LSTM [3] network, we find that although it works fine on QQP dataset[1], it fails at various places on domain-specific search queries. As our use-case is specific to the financial domain, we use the transfer learning approach to enable the model to generalize on the domain-specific vocabulary. We add our internal financial domain-specific data to the training process and tune it to improve the precision of the positive class. Furthermore, we calculate the similarity score using Siamese [6] LSTM [3] network after normalizing the search queries. Even then the model's performance does not improve significantly. Analysing the misclassified instances, we comprehend that the failure is due to incomplete, misspelt and grammatically incorrect queries.

Some examples are as follows:

"select sell all shares instead" and "do i choose just sell or sell all shares"

"can you explain to me what a mutual fund is" and "and i also want to know about mutual funds"

*"Fractional Trading" and "What does Factnol Trding means"*

To tackle these issues, we experiment with different similarity scores like fuzzy-match score and cosine similarity based on average embedding. We combine these four scores (un-normalized,

**Algorithm 1:** Calculate customized Fuzzy Intersection Ratio between two questions

**Input:** question1 (input query), question2 (question from QA corpus), threshold1, threshold2, vocabulary
**Output:** Fuzzy Intersection Ratio (i.e. $fuzzy\_set\_ratio$)

1: $s1 \leftarrow set(question1.tokenize())$
2: $s2 \leftarrow set(question2.tokenize())$
3: $overlap \leftarrow s1 \cap s2$
4: **if** $length(overlap) = 0$ **then**
5:    $fuzzy\_set\_ratio \leftarrow 0$
6: **else**
7:    $exclusive\_s1 \leftarrow s1 - s2$
8:    $exclusive\_s2 \leftarrow s2 - s1$
9:    $rem\_s1 \leftarrow \{\}$
10:   $rem\_s2 \leftarrow \{\}$
11:   $partial\_overlap \leftarrow 0.0$
12:   **for** each $word1$ in $exclusive\_s1$ **do**
13:     **for** each $word2$ in $exclusive\_s2$ **do**
14:       $dist \leftarrow norm\_levenshtein\_distance(word1, word2)$
15:       $syntactic\_similarity \leftarrow 1.0 - dist$
16:       **if** $syntactic\_similarity \geq threshold1$ **then**
17:         $partial\_overlap \leftarrow partial\_overlap + syntactic\_similarity$
18:         $rem\_s1 \leftarrow rem\_s1 \cup \{word1\}$
19:         $rem\_s2 \leftarrow rem\_s2 \cup \{word2\}$
20:         **break**
21:       **else if** $word1$ in vocabulary & $word2$ in vocabulary **then**
22:         $semantic\_score \leftarrow norm\_cosine\_similarity(word1, word2)$
23:         **if** $semantic\_score \geq threshold2$ **then**
24:           $partial\_overlap \leftarrow partial\_overlap + semantic\_score$
25:           $rem\_s1 \leftarrow rem\_s1 \cup \{word1\}$
26:           $rem\_s2 \leftarrow rem\_s2 \cup \{word2\}$
27:         **else**
28:           skip to next iteration
29:         **end if**
30:       **else**
31:         skip to next iteration
32:       **end if**
33:     **end for**
34:   **end for**
35:   $exclusive\_s1 \leftarrow exclusive\_s1 - rem\_s1$
36:   $exclusive\_s2 \leftarrow exclusive\_s2 - rem\_s2$
37:   $lo \leftarrow length(overlap)$
38:   $ls1 \leftarrow length(remaining\_s1)$
39:   $le1 \leftarrow length(exclusive\_s1)$
40:   $le2 \leftarrow length(exclusive\_s2)$
41:   $fuzzy\_set\_ratio \leftarrow \frac{lo+partial\_overlap}{lo+ls1+le1+le2}$
42: **end if**

normalized, average embedding and fuzzy-matching based) using a Support Vector Machine [19] based meta-classifier (SVC). When compared to the best performing SOTA model (RoBERTa mentioned in Table : I), M1 (SVC on 4 scores) shows an improvement of 0.09 and 0.25% on F1-score (un-weighted, macro average) and Accuracy respectively on our internal test set. For improving its performance further, we try to combine one more score. It is the cosine similarity score between sentence embeddings [7] of questions generated using pre-trained transformer-based models like XLNet [1], BERT [11], DistillBERT [16] and RoBERTa [17]. The improvements in performances after including the scores generated using these transformer-based embeddings is comparable. However, we decide to use the RoBERTa [17] based score as the fifth input to the SVC model as it performs the best on our internal test dataset. After performing a grid search, we decide to use 0.7 as the threshold for classification as it gives the best precision value for the positive class. All the training iterations have been performed on a Nvidia DGX GPU environment. It has 32 Nvidia Tesla V100 GPUs. It is made up of over 160,000 CUDA Cores and over 20,000 Tensor Cores.

---

[5] https://cims.nyu.edu/~sbowman/multinli/

We present the results of the SOTA models (like XLNet [1], BERT [11], DistillBERT [16] and RoBERTa [17] ) as separate rows in Table I. Its columns represent F1 score (unweighted, macro averaged) and Accuracy (as a percentage) for QQP[1] and internal test dataset. The cells under QQP SOTA is populated by referring to GLUE[3] [21] leader-board and DistillBERT [16] paper. The numbers (denoted by *) for internal test (Int. Test) domain specific set have been obtained using cosine similarity of sentence embeddings [7] generated using the models mentioned in the corresponding rows. We use a threshold of 0.7 which is obtained empirically. It proves the fact that off-the-self SOTA models fail miserably when applied as it is on domain specific data.

The results we obtain using the models we experiment with are mentioned in Table II. It consists of two main columns which are divided into two sub-columns each.

Table I:
PERFORMANCE OF SOTA MODELS ON QQP AND INTERNAL TEST SET

| Model | QQP SOTA | | Int. Test | |
|---|---|---|---|---|
| | F1 | Acc% | F1* | Acc*% |
| **XLNet Ensemble** (Multi-task ensemble) | 0.747 | 90.40 | 0.53 | 70.96 |
| **BERT** (24-layers, 16-heads, 1024-hidden) | 0.721 | 89.30 | 0.58 | 72.05 |
| **DistillBERT** | - | 88.50 | 0.61 | 72.47 |
| **RoBERTa** (Multi-task ensemble) | 0.743 | 90.20 | 0.65 | 74.56 |

Table II:
PERFORMANCE OF THE MODELS DEVELOPED BY US
M1 = SVC (4 scores), M2 = SVC (4 scores + XLNet), M3 = SVC (4 scores + BERT), M4 = SVC (4 Scores + DistillBERT), M5 = SVC (4 Scores + RoBERTa)

| Model | QQP Test | | Int. Test | |
|---|---|---|---|---|
| | F1 | Acc% | F1 | Acc% |
| M1 | 0.80 | 81.17 | 0.74 | 74.81 |
| M2 | 0.80 | 81.29 | 0.74 | 75.23 |
| M3 | 0.80 | 81.36 | 0.74 | 75.23 |
| M4 | 0.80 | 81.22 | 0.74 | 75.15 |
| M5 | 0.80 | 81.18 | 0.75 | 75.65 |

These two main columns state the performances of our models on the holdout QQP1 dataset and internal test set respectively. The sub-columns represent F1 score (unweighted, macro averaged) and Accuracy (as a percentage) respectively. The rows represent the models we tried out. M1 is the SVC model which uses four scores (un-normalized, normalized, average embedding and fuzzy-match). Details relating to these scores have been described in section 4. Models M2, M3, M4 and M5 represent SVC models trained using five scores where four scores are the same as before. The fifth score is the cosine similarity score between sentence embedding [7] of each of the two input question. This sentence embedding [7] is generated either using XLNet (M2) or BERT (M3) or DistillBERT (M4) or RoBERTa (M5). To create these sentence embeddings [7], the models mentioned here have been trained on SNLI[2] and MultiNLI[5] dataset. In each case, mean tokens pooling has been used. More details relating to the creation of sentence embedding from a text corpus can be found here[6].

## 6 DISCUSSION

In this section, we discuss the results we presented in the last section. We also narrate how we have put this system into production.

### 6.1 Discussion

On carefully examining Table I and II, we see that variants of transformer model work very well on the QQP dataset[1] but not on our domain specific internal test set. Comparing the performance of SOTA models with the models presented in this paper we observe that, although SOTA models outperform all the proposed models on QQP dataset[1] when evaluating on accuracy the presented models perform better in terms of F1 score. Models presented in the paper outperform SOTA models both on accuracy and F1-score for our internal test set. This leads us to conclude that trade-off exists between the performances when measured against QQP dataset[1] versus our internal dataset. Thus, we decide to compromise a bit with the performance on the QQP dataset[1] to get better performance for our specific use-case, that is for search queries which appear in the financial domain.

---
[6]https://github.com/UKPLab/sentence-transformers/blob/master/docs/pretrained-models/nli-models.md

SVC with RoBERTa (M5) having a F1 score of 0.75 and accuracy of 75.65% performs the best for our internal test dataset. Thus, we select SVC with RoBERTa (M5) model as our final model and move forward to put it in production. Its precision for the positive class is 0.93, accuracy is 75.65% and F1 score (un-weighted, macro average) is 0.75 on the internal test set. It captures intricacies present in the financial search queries as well as general search queries. This ensures that even if queries differ semantically over time, SVC with RoBERTa model (M5) will be able to comprehend it. Some instances are as follows:

*"is there anyway to contribute to my previous employer's fund"* and *"is it still possible to contribute to my previous employer sponsored plan"* *"what do we need to do to transfer these roth accounts"* and *"how to move roth accounts"*

Moreover, we compare the final model's (M5 : SVC with RoBERTa) performance to that of the SOTA models (Table I). We conclude that on QQP dataset[1], its accuracy is less than SOTA models by 7% to 9%, but its F1 score is more than these models by at least 0.079. Although the test set for the SOTA models is different from that of ours, we still move ahead with the comparison as the latter is a random holdout set (having 80,872 question pairs) created from the entire QQP dataset[1] (having 404,290 question pairs).

## 6.2 Implementation

The proposed framework has been implemented using Python. We have used Keras[7] with TensorFlow[8] at the backend for modelling and Spacy[4] for pre-processing (Parts of Speech tagging and Lemmatization) the input text. This framework is modular, and it is capable of retrieving the most relevant QA pair for the user query in real-time. In production, we aim to match a query coming from a user to one of the questions present in the curated QA corpus which is in production.

This means that the incoming query has to be compared with all the questions in the QA corpus which is computationally expensive. Furthermore, the time required for generating the similarity scores tends to increase when the size of QA corpus increases. Thus, this process is not usable here. To enhance performance, we divide it into two parts:

**Embedding creation:** We pre-calculate different kinds of embeddings (normalized, un-normalized, average and sentence embeddings using RoBERTa) for every question present in the QA corpus only once and store them in a Numpy[9] array. Only embeddings for the incoming question is calculated in production.

---
[7] https://keras.io/
[8] https://www.tensorflow.org/

We pass these embedding (both normalized and un-normalized) through one of the arms of the Siamese network [6] and use the corresponding output from the pre-final layer.

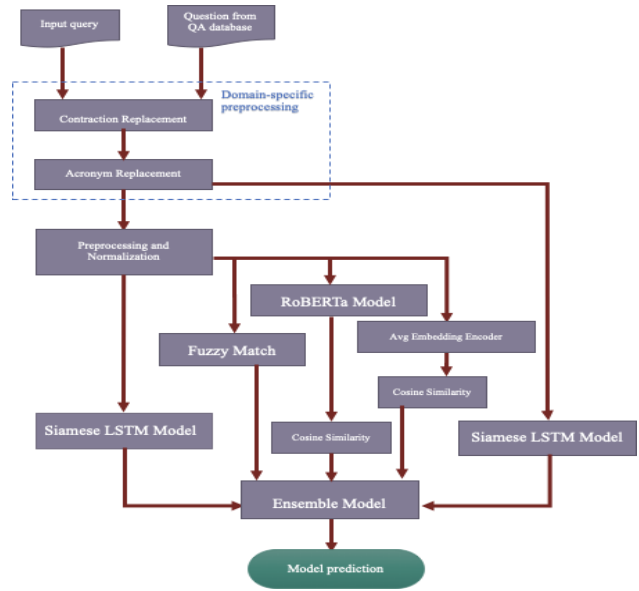

Figure 2: Flowchart

**Similarity calculation:** Once we have the output from the pre-final layer in hand, we calculate the similarity scores (normalized and un-normalized). This is done using vectorized operations with the pre-computed embeddings. Similar vectorized operations are performed to obtain similarity scores using word2vec (averaged) [15] and RoBERTa [17]. These similarity scores along with the fuzzy-match score are passed to the SVC classifier [19] to predict the final similarity score. Finally, the question having the highest similarity score along with its corresponding answer is returned. This whole framework is designed to enhance the users' search experience. It can easily be generalized to other domains as well without compromising much on the performance.

## 7 CONCLUSION AND FUTURE WORKS

After comparing performances of all the models, we conclude that our final model (M5) does 0.1 and 1% better than best SOTA models in terms of F1 and accuracy respectively. Performance comparison and error analysis on internal test set points to the fact that the proposed model/framework is better suited for search setting and would perform better for domain specific applications. In future, we want to test how Hierarchical Attention Networks [14] perform if they are used to combine words to represent sentences. We would like to try out different kinds of matching mentioned in [12] in future. Moreover, we would like to incorporate Named

---
[9] https://numpy.org/

Entity identifier and BERT embedding [11] which has been customized specifically for queries searched on the web and mobile platforms of Fidelity. In one of the papers, recently released by Google [18], Raffel et al. narrate how they used a single model referred to as Text-to-Text Transfer Transformer (T5) to accomplish a wide variety of NLP tasks like question answering, translation, sentence similarity calculation and so on. We are considering to explore this model's method of calculating semantic textual similarity. We shall try replacing the word2vec embedding layer [15] in the model proposed by other embeddings like XLNet [1], BERT [11] and so on. This may improve the model's performance further. Finally, we shall be evaluating this model on stack-exchange data[10]. This will enable us to benchmark our model better.

## ACKNOWLEDGMENTS

We thank Chaitra Vishwanatha Hegde for suggesting us to use SVC for combining the scores mentioned in section 4.

---

[10] https://archive.org/details/stackexchange